\documentclass[journal]{IEEEtran}

\usepackage{indentfirst}
\usepackage{bm}
\usepackage{amsfonts}
\usepackage{times}
\usepackage{multirow}
\usepackage{booktabs}
\usepackage{graphicx}

\usepackage{amsmath}
\usepackage[export]{adjustbox} 

\usepackage{diagbox}
\usepackage{multirow}
\usepackage{booktabs}

\usepackage{algorithm}
\usepackage{algorithmic}

\usepackage[pagebackref=true,breaklinks=true,letterpaper=true,colorlinks,bookmarks=false]{hyperref}

\begin{document}

\title{Blind Image Quality Assessment via Transformer Predicted Error Map and Perceptual Quality Token}

\author{Jinsong Shi,~ Pan Gao, and Aljosa Smolic \\ 
\thanks{J. Shi and P. Gao are with College of Computer Science and Technology, Nanjing University of Aeronautics and Astronautics, China. A. Smolic is with Lucerne University of Applied Sciences and Arts, Lucerne, Switzerland. }
}

\maketitle

\begin{abstract}
Image quality assessment is a fundamental problem in the field of image processing, and due to the lack of reference images in most practical scenarios, no-reference image quality assessment (NR-IQA),  has gained increasing attention recently. With the  development of deep learning technology, many deep neural network-based NR-IQA methods have been developed, which try to learn the image quality based on the understanding of database information. Currently, Transformer has achieved remarkable progress in various vision tasks. Since the characteristics of the  attention mechanism in Transformer fit the global perceptual impact of artifacts perceived by a human, Transformer is thus well suited for image quality assessment tasks. In this paper, we propose a Transformer based NR-IQA model using a predicted objective error map and perceptual quality token. Specifically,  we firstly generate the predicted error map by pre-training one model consisting of a Transformer encoder and decoder, in which the objective difference between the distorted and the reference images is used as supervision. Then, we freeze the parameters of the pre-trained model and design another branch using the vision Transformer to extract the perceptual quality token for feature fusion with the predicted error map. Finally, the fused features are regressed to the final image quality score. Extensive experiments have shown that our proposed method outperforms the current state-of-the-art in both authentic and synthetic image databases. Moreover, the attentional map extracted by the perceptual quality token also does conform to the characteristics of the human visual system.

\end{abstract}


\IEEEpeerreviewmaketitle

\section{Introduction}
With the popularity of the Internet and the rapid development of social networks, a large number of images are generated from them. As the quality of images directly affects people's viewing experience,  the assessment of image quality is extremely important. In addition, in the fields of image compression \cite{mier2021deep} and image enhancement \cite{chen2014quality}, a good IQA method will also become an indicator to measure the performance of different image processing algorithms. Human evaluation is time-consuming and laborious, so an effective objective IQA method is especially important.

IQA algorithms can be typically classified into three categories according to the presence or absence of a reference image: full-reference (FR) \cite{kim2017deep,Cartoon_quality}, reduced-reference (RR) \cite{golestaneh2016reduced}, and no-reference (NR) \cite{ye2012unsupervised}. Although favorable results have been achieved for the FR and RR methods, the NR-IQA method has a wider range of applications in the real-world and has been a prevalent area of IQA research. The NR-IQA is also the most difficult and challenging one in the IQA tasks since it completely lacks of the information from the reference image.

NR-IQA can generally be divided into distortion-specific methods (\textit{e.g.}, blur, JPEG) \cite{li2015no,liu2009no} and general-purpose methods \cite{mittal2012no,saad2012blind,xue2014blind,ye2012unsupervised,gao2013universal}. In the distortion-specific approach, only the characteristic information of the considered distortion needs to be extracted, and this approach has achieved good results so far, basically similar to the human subjective evaluation results. However, this approach also has considerable limitations, because there are various kinds of image distortions and many unknown distortions. It will not work well when facing a new type of distortion, and thus does not have good generalization performance. In comparison to IQA methods for specific distortions, generic methods focus on extracting generalized image distortion information through hand-crafted \cite{saad2012blind} or learned features \cite{ye2012unsupervised}, which can be used to evaluate images with various specific distortion types, as well as generalize to mixed distortion types and unknown image distortions. Therefore, the main attention of NR-IQA research is devoted to generic methods.

Most of the present general-based methods perform better on the traditional synthetic distortion databases LIVE \cite{sheikh2005live}, TID2013 \cite{ponomarenko2015image}, and CSIQ \cite{larson2010most}, and they perform poorly on the real distortion databases LIVE Challenge \cite{ghadiyaram2015massive} and KonIQ-10k \cite{hosu2020koniq}. The reason is that there are fewer reference images in the synthetic distortion database, usually, no more than 30, of which LIVE contains 29, CSIQ contains 30, and TID2013 contains 25. On the real distortion database, LIVE Challenge contains 1162 reference images, and KonIQ-10k has 10,073. More reference images imply more distortion types, which require a higher generalization performance of the NR-IQA model. In addition, authentic image distortion is complex and diverse. Because there do not exist real reference images in the real world, such images are difficult to be evaluated even for normal humans. Some of these distorted images that involve aesthetic aspects  may only be felt by humans to be more in line with the real aesthetics \cite{wang2009mean,li2018has}, while models will be difficult to judge.

Currently, NR-IQA has been considered as a linear regression problem, \textit{i.e.}, an IQA model needs to be designed with distorted images as input and corresponding scores as output. The output is usually MOS/DMOS values. Such a pattern of design will lead to a model lacking human subjective HVS information, and therefore the prediction accuracy of the model is limited. To tackle this problem, we propose a novel NR-IQA method based on objective distortion maps and the human visual saliency effect. The contributions of our work are summarized below.

\begin{itemize}
    \item 
    We propose a novel NR-IQA method that leverages Transformer's self-attention mechanism and CNN inductive bias. Unlike existing methods, our proposed model not only predicts accurately on synthetic databases but also performs well on authentic databases.
    \item 
    In order to enable the model to learn the distortion information in the image accurately, we train the model to learn the objective error map by using the difference between the distorted image and the reference image as the supervised information. In addition, we fuse the predicted distortion map information with the perceptual quality token learned in the Transformer. The regressed distortion image quality score is more in line with human visual characteristics.
    \item
    We have conducted extensive experiments on the current major IQA databases. The experimental results show that our proposed model achieves the state-of-the-art, and the generated predicted error maps are consistent with HVS characteristics.
\end{itemize}

Our implementation code and pre-trained models are available at the link: \href{https://github.com/Srache/TempQT}{https://github.com/Srache/TempQT}.

The rest of this paper is organized as follows. Section II reviews the related works. Our proposed transformer-based NR-IQA model is presented in Section III. Experiments are conducted in Section IV, followed by conclusion remarks in Section V.

\section{Related Work}
Before the emergence of deep neural network methods, traditional NR-IQA methods could be divided into hand-crafted feature modeled \cite{moorthy2010two,saad2012blind,mittal2012no} and learning-based \cite{ye2012unsupervised,zhang2015som} categories. In the first category, NSS is a commonly used hand-crafted feature, which means that the visual feature information of distortion-free images follows a certain distribution rule. Since different types and degrees of distortion will have an impact on this rule, different NR-IQA methods can be thus designed according to this characteristics. Moorthy \textit{et al}. \cite{moorthy2010two} used discrete wavelet transform (DWT) to extract NSS features for evaluating reference-free images. Saad \textit{et al}. \cite{saad2012blind} used statistical features of discrete cosine transform (DCT) to evaluate image quality. Mittal \textit{et al}. \cite{mittal2012no} proposed to use NSS features in the spatial domain to construct an image quality assessment model and achieved good performance. On the other hand, learning-based approaches such as using dictionary learning method in machine learning, Ye \textit{et al}. \cite{ye2012unsupervised} proposed a NR-IQA algorithm based on dictionary learning to obtain image visual perceptual features by constructing code books through K-means, and then used support vector regression (SVR) model to predict the subjective quality score of distorted images. Zhang \textit{et al}. \cite{zhang2015som} combined semantic-level features affecting the human visual system (HVS) with local features for image quality estimation. Although the aforementioned methods based on hand-crafted features and automatic learning perform relatively well on some synthetic databases, the results on real databases are less impressive.

\noindent\textbf{Deep learning for NR-IQA.} Different from traditional hand-crafted features, NR-IQA models based on deep learning \cite{kang2014convolutional,kim2018deep,yan2018two,lin2018hallucinated,zhu2020metaiqa,su2020blindly} can learn the end-to-end mapping relationship between image and image quality and perform significantly better than traditional machine learning-based models. In the early time, most deep learning-based NR-IQA approaches focus on the architecture  design of using convolutional neural networks (CNNs). Kang \textit{et al}. \cite{kang2014convolutional} introduced CNN into the NR-IQA model design and used simple linear regression to predict quality scores. Kim \textit{et al}. \cite{kim2018deep} divided the training of the NR-IQA model into two stages, with the first stage training a model for obtaining a local quality map and the second stage fine-tuning the model and predicting the human subjective evaluation scores. Yan \textit{et al}. \cite{yan2018two} proposed an NR-IQA model based on a dual-flow CNN structure, using two sub-networks with the same structure to extract the distortion map and the corresponding gradient map features separately. Lin \textit{et al}. \cite{lin2018hallucinated} proposed an NR-IQA model based on generative adversarial networks (GANs), where they first generated the hallucinated reference image to compensate for the absence of the real reference and then paired the hallucinated reference information with the distorted image to estimate the quality score. Zhu \textit{et al}. \cite{zhu2020metaiqa} proposed a model that uses meta-learning to learn prior knowledge shared between images of different distortion types. The NR-IQA method proposed by Su \textit{et al}. \cite{su2020blindly} extracts content features at different scales from a deep model and brings them together to predict image quality.

\noindent\textbf{Transformers for NR-IQA.} The significant success of CNNs in computer vision is largely facilitated by locality and spatial invariance, but CNNs are less focused on the long-term dependence on the images. IQA can be considered essentially as a recognition task, \textit{i.e}., recognizing the quality level of an image, and therefore needs to be assessed by combining local and global information about the image. Transformers \cite{vaswani2017attention}, which were first advanced in the field of NLP, completely remove the CNN structure and can naturally obtain the long-term dependence information of sequences due to the specialized attention mechanism. In the past two years, Transformers have been used with great success in various tasks in the field of computer vision \cite{dosovitskiy2020image,carion2020end,zheng2021rethinking,liu2021swin}, and they have also been applied in the field of NR-IQA \cite{golestaneh2022no}. Golestaneh \textit{et al}. \cite{golestaneh2022no} proposed a hybrid NR-IQA model based on CNN and Transformers to design the ranking loss among distorted images and proposed a consistency loss of flip invariance of distorted images, which has yielded remarkably good results on both synthetic and authentic databases.

\begin{figure*}
    \centering
    \includegraphics[scale=0.7]{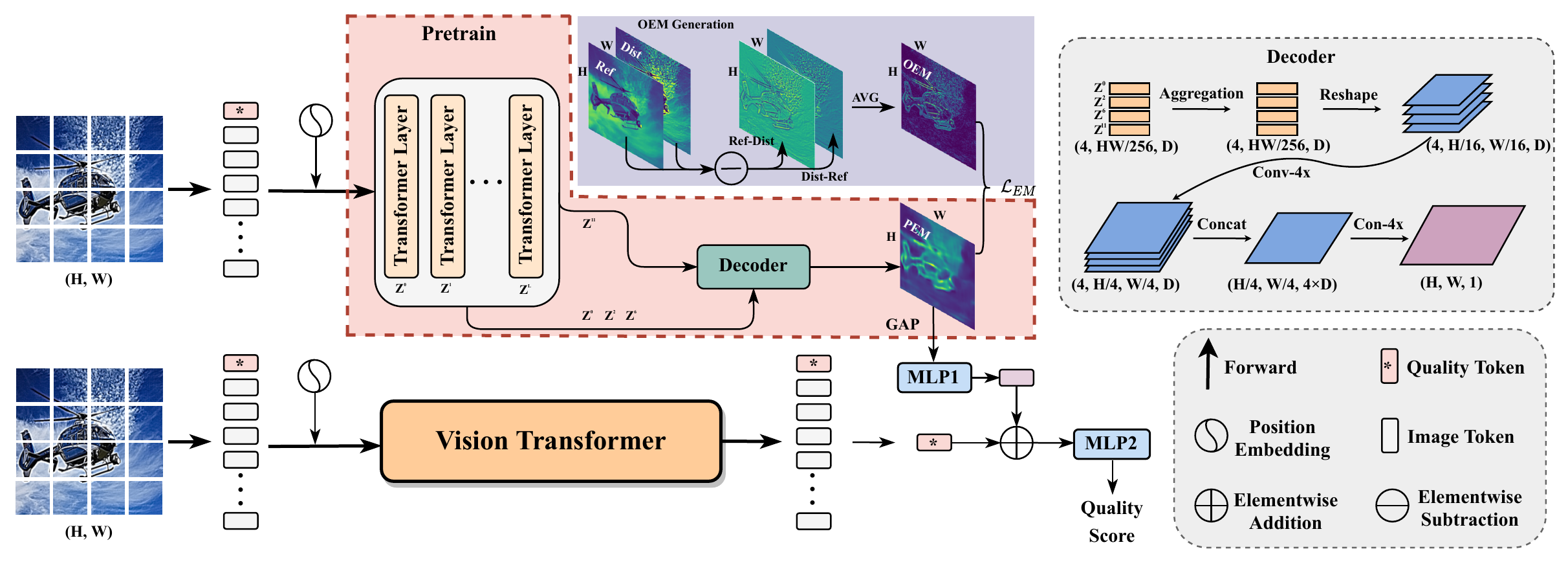}
    \caption{The overall framework of our approach for no-reference image quality assessment. Ref and Dist represent reference image and distorted image, respectively. Ref-Dist means using the reference to subtract the distorted image, and vice versa. OEM denotes ground-truth objective error map, and PEM is the predicted error map. The Ref and Dist are inputted in grayscale space, which are shown here using viridis color.}
    \label{fig:fig1}
\end{figure*}

\section{Proposed Method}

In this section, we detail our proposed model, which is an NR-IQA approach based on Transformer predicted Error Map and Perceptual Quality Token, namely \textit{\textbf{TempQT}}. The architecture of our network is shown in Figure \textcolor{red}{\ref{fig:fig1}}, which is composed of two steps, \textit{i.e.}, the objective error map model pre-training and  image score prediction. In the first step, we leverage the difference between the distorted image and the reference image as Ground-Truth to train the Transformer model to generate an error map, where the size of the prediction error map is the same as the input model. In the second step, we first freeze the weights of the pre-trained model, and then train another transformer-based quality assessment model to produce a perceptual quality token, which is then  fused with the predicted error map from the pre-trained model.  Finally, the fused information is used for the final quality score prediction of the distorted image.
\subsection{Objective error map prediction model}
In our framework, we first pre-train a model to generate the objective error map for the input image. In this model, we employ the original transformer as the backbone and design a decoder that aggregates the patch embedding of different layers in the transformer for error map prediction. Note that, during training, this model requires the ground truth error map as supervision. In other words, we need the reference image for training this model. However, once the error map prediction model is trained, we no longer need the reference image. In our subsequent quality evaluation module, the pre-trained error map model can be used to infer a plausible error map directly without needing reference image. Therefore, our quality assessment model is blind. 

\begin{figure}
    \centering
    \includegraphics[scale=0.7]{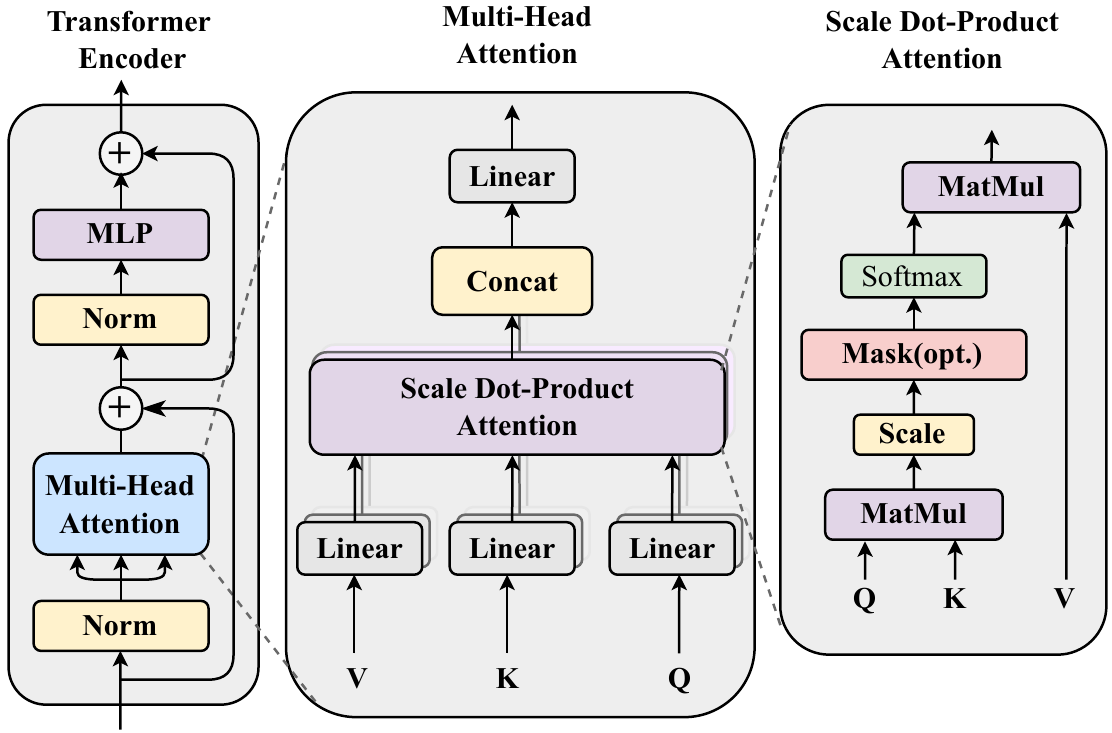}
    \caption{Illustration of the architecture of the Transformer Encoder.}
    \label{fig:fig2}
\end{figure}

\noindent\textbf{Transformer.} We choose the ViT \cite{dosovitskiy2020image} as the vision Transformer backbone. Given a 2D image $\mathbf{x} \in \mathbb{R}^{H \times W \times C}$, we reshape the image into 2D patches $\mathbf{x}_{p} \in \mathbb{R}^{N \times P \times P \times C}$, where ($H$,$W$) is the resolution of the original image, $C$ is the number of channels, $P$ is the size of the patch, and $N$ is the number of patches ($N=HW/P^2$). 
Since Transformer uses a constant size D-dimensional latent vector as the feature representation of the sequence at each layer, we flatten 2D patches and map to $D$ dimensions by a linear projection whose parameters can be learned. 
In the instance of ViT-b16, where $D$ is 768, if $P$ is set to 16, a $224\times224$ input image $\mathbf{x}$ will eventually map into a sequence of patch embedding of dimension $196\times 786$, and in this case, $D$ equals  $P\times P\times C$. 
In order to encode the image spatial information, we add a learnable position embedding $p_{i}$ for each patch. So the final input sequence $Z^{0}$ = $\left \{ s_{1}+p_{1}, s_{2}+p_{2}, ..., s_{N}+p_{N} \right \} $, where $s_{i}$ represents patch embedding. The encode of transformer is composed of L-layer multi head self attention (MHSA) and multi layer perceptron (MLP) blocks. At layer $l$, the input of self-attention consists of three parts: query, key and value. Assume $Q$ as query, $Q=Z^{l-1}W_Q$, where $Z^{l-1}$ denotes the output of the previous layer, $W_Q\in \mathbb{R}^{D\times d}$ is the the learnable parameters of linear projection layer and $d$ is the dimension of query. Key and value can be calculated similarly. With query, key and value, the Self-attention(SA) is calculated as follows: 
\begin{equation}
SA(Z^{l-1})=Z^{l-1}+softmax\left(\frac{Q\times K^T}{\sqrt{d}}\right)\times V 
\end{equation}
MHSA is an extension with $h$ independent SA operations and projects their concatenated outputs using: $MHSA(Z^{l-1})=Concat(SA_1(Z^{l-1}),SA_1(Z^{l-1}),...,SA_h(Z^{l-1}))W$, where $W\in \mathbb{R}^{{h\times d\times D}}$, and $d$ is typically set to $D/h$. The output of the MHSA will go through the MLP layer and be added back via the residual connection.  The final output at layer $l$ is: $$Z^{l} = MHSA(LN(Z^{l-1})) + MLP(MHSA(LN(Z^{l-1})))$$ We denote $\left \{ Z^{1},Z^{2},...,Z^{L} \right \} $ as the output features of Transformer layers.  The overall structure and components of ViT are shown in Figure \textcolor{red}{\ref{fig:fig2}}.

\noindent\textbf{Decoder.}
The goal of decoder is to generate objective error maps corresponding to distorted images. Considering the characteristics of Multi-Level encoder design of transformer, the output results of the specified encoder layers are first selected as representative features, and in this paper we select layers $Z^{0}$, $Z^{2}$, $Z^{6}$ and $Z^{11}$. To enhance the interaction information among different layers, we do element-wise summation for the features of different layers from top to bottom (\textit{i.e.} $\hat Z^{2} = Z^{2} + Z^{0}, \hat Z^{6} = Z^{6} + \hat Z^{2}, \hat Z^{11} = Z^{11} + \hat Z^{6}$). We then  reshape the output of the aggregated layers to turn the 2D sequence ($\frac{HW}{256} \times D$) into a 3D feature map ($\frac{H}{16}\times \frac{W}{16}\times D$). Then, we use a convolution of 3$\times$3  for the feature maps, and upsample the output of each layer by a factor of 4 using bilinear interpolation. Finally, we concatenate the upsampled 4-layer output along the dimension of the channel, and then use bilinear interpolation for 4$\times$ upsampling to restore the feature map to the resolution of the original error map ($H\times W\times 1$).
Note that, since each cascaded transformer layer may capture different types of features from the input image, which layers are used in decoder may have an impact on error map generation. We will ablate the selection of layers in Section IV. 

\noindent\textbf{Objective Error Map.} We calculate the Objective Error Map (OEM) based on the difference between the distorted image and the reference image and use it as supervised information to train the Transformer encoder and decoder. OEM is calculated as follows:
\begin{equation}
\label{eq1}
    OEM = \frac{\left | Dist-Ref \right | + \left | Ref-Dist \right | }{2} 
\end{equation}
where $Dist$ denotes the distorted image, and $Ref$ denotes the reference image. $Dist$ and $Ref$ are both grayscale images.

The loss of the objective error map at pre-training is defined as follows:
\begin{equation}
\label{eq2}
\mathcal{L}_{EM}=\left \| PEM-OEM \right \|_2^2+\lambda \left \|Ref'-Ref \right \|_2^2
\end{equation}
Where $PEM$ denotes the Predicted Error Map by the model, $\lambda$ represents the balance factor, and $Ref^{'} = Dist-OEM$.

\subsection{Perceputal Quality Token (PQT) Generation}
After pretraining the transformer to get the objective error map, we employ another vision transformer to obtain the perceptual importance of each patch in the input image.  
As the class token in the original Vision Transformer is used for image classification, it contains mainly the information about the object category in the image. When applied to the IQA task, we use  a perceptual quality token instead of the class token, which is also learnable. Denote by $Q_{PQT}^{l}$ the query quality token at layer $l$, and its dimension is $1 \times d$. The output quality token calculated using self-attention  can be expressed as follows:
\begin{equation}
\begin{aligned}
    Z_{PQT}^{l} &= softmax\left(\frac{Q_{PQT}^{l}\times (K^l)^T}{\sqrt{d}}\right)\times V^l \\
    &=A^{l}_{PQT}\times V^l
\end{aligned}
\end{equation}
where $K^l$ and $V^l$  are the key and value embedding at layer $l$, respectively. Both have the dimension of $N \times d$. $A^{l}_{PQT}$ represents the attention vector of the PQT token, which is obtained by dot-product of the PQT token with all other patch tokens followed by a softmax operation. The PQT token at each layer $Z_{PQT}^{l}$ is obtained by multiplying $A^{l}_{PQT}$ with patch embedding $V^l$. Since $A^{l}_{PQT}$ indicates how much attention of the PQT is paid on each patch embedding, $Z_{PQT}^{l}$ contains the perceptual impact of each patch token on the evaluated image. Thus, in the next subsection, we will use the learned PQT for feature aggregation for image quality prediction.

To verify the effectiveness of the PQT, we will extract the overall Attention Map (AM) learned by the transformer for visualization, and the AM is calculated as follows. Firstly, as shown above, the attention vector of PQT at each layer is presented as a vector of dimension $N$, i.e., $A^{l}_{PQT}$. If each layer has $h$ heads, the attention vector of each layer is updated as the average of attention vectors from $h$ heads. 
Then, the overall attention vector of PQT $A_{PQT}$ is calculated by averaging the attention vector of all layers in the transformer. 
In order to get the perceptual attention map of the PQT having the same spatial size as original image, the  attention vector is first reshaped into a map of dimension $\sqrt{N}\times \sqrt{N}$, and then mapped to the original size using interpolation methods. The final Attention Map is of size $H\times W$. The generated perceptual attention map will be shown in Section IV. As will be seen, the attention map focuses on the  perceptually distorted part, which is basically the same as the perceptual region of humans in evaluating image quality. Therefore, perceptual attention aware PQT is beneficial for evaluating image quality.

\subsection{Feature fusion and quality score prediction}
When evaluating a distorted image, humans not only are sensitive to the distortion information of the distorted image, but also, they may tend to have different perceptual experiences when facing a same amount of distortions but occurred at different regions in the image.  
 Therefore, it is unreasonable to generate only an error map by the model for image quality evaluation directly, which does not take into account the perceptual  information of the distorted image perceived by humans. Therefore, we design a two-branch structure, where one branch is used to extract the distortion information of the image, and the other branch is used to produce the perceptual attention related to human visual system mechanism. These two branches are finally fused together for quality score prediction.

To perform feature fusion on these two branches, we firstly use a global average pooling (GAP) operation on the PEM output of the pre-trained model, followed by using the Multilayer Perceptron1 (MLP1) to change the dimension of the global vector to $D$. This gives us the objective distortion information vector $V_{PEM}$ of the image. The calculation of $V_{PEM}$ can be represented  as follows:
\begin{equation}
    V_{PEM} = FC(GAP(PEM))
\end{equation}

Then, we extract the perceptual quality token $Z_{PQT}^{L}$ from the last layer of the  transformer in the second branch. In the training process for  image score prediction, the parameters of the pre-trained network model for generating PEM are frozen, and only the parameters of the second branch need to update. Eventually, we do element summation for $V_{PEM}$ and $Z_{PQT}^{L}$ to get the fused image quality features, which are then regressed by Multilayer Perceptron2 (MLP2) into the final  subjective quality score. This process can be formulated as:
\begin{equation}
    Preds = FC(PReLU(FC((V_{PEM}+ Z_{PQT}^{L}))))
\end{equation}
In each batch of images, we train our quality prediction model by minimizing the regression loss as follows
\begin{equation}
\label{eq6}
    \mathcal{L}_{Q}=\frac{1}{N} \sum_{i=1}^{N}\left\| preds_{i}-y_{i}\right\|_1
\end{equation}
where $N$ denotes the batch size, $preds_{i}$ denotes the  image quality score predicted by the model for the $i^{th}$ image, and $y_{i}$ denotes the corresponding objective quality score. The whole procedure of the proposed blind image quality evaluation method is summarized in Algorithm \textcolor{red}{\ref{alg1}}.

\begin{algorithm}
	\textbf{Require:} Distorted image (Dist), Reference images (Ref), Ground Truth scores, Learning rate $\alpha$ and $\beta$
	
	\textbf{Output:} Prediction scores
	\caption{PEM and PQT based NR-IQA}
	\label{alg1}
	\begin{algorithmic}[1]
	    \STATE Loading pre-training model parameters $\theta$ from ViT-B/16
		\STATE /* PEM model Pre-training */
		\STATE /* input: Dist, Ref; output: PEM*/
		\STATE $\textbf{Subroutine\_Sub:}$ \textbf{~\{}
        \FOR{$\textit{iteration} ~i = 1,2,...$}
        \STATE Compute $\theta_{PEM}^{i}=Adam(\mathcal{L}_{EM}, \theta)$;
        \STATE Update $\theta_{PEM} \gets \theta - \alpha (\theta - \theta_{PEM}^{i})$;
        \ENDFOR ~\textbf{\}}
        \STATE /* Quality score prediction */
		\STATE /* input: Dist, PEM, Ref; output: prediction score */    
        \STATE  $\textbf{Main routine:}$ \textbf{~\{} 
		\FOR{$\textit{iteration} ~i = 1,2,...$}
		\STATE /* Freeze model parameters $\theta_{Pre}$ */
		\STATE Call subroutine\_Sub to output PEM;
		\STATE Compute $\theta_{Q}^{i}=Adam(\mathcal{L}_{Q}, \theta)$;
        \STATE Update $\theta_{Q} \gets \theta - \beta (\theta - \theta_{Q}^{i})$;
		\ENDFOR ~\textbf{\}}
	\end{algorithmic}  
\end{algorithm}



\section{Experiments}
\subsection{Datasets}

We evaluate the performance of our method using major IQA datasets containing three synthetic databases LIVE \cite{sheikh2006statistical}, CSIQ \cite{larson2010most}, TID2013 \cite{ponomarenko2015image}, KADID-10K \cite{lin2019kadid} and two authentic databases LIVEC \cite{ghadiyaram2015massive}, KonIQ-10K \cite{hosu2020koniq}. The KADID-10K database is mainly used for objective error map training and is not involved in performance evaluation. Table \textcolor{red}{\ref{tab:1}} lists the summary information for each database.

The commonly used observer ratings of the images are expressed by Mean opinion score (MOS) and Differential Mean opinion score (DMOS), where larger MOS values indicate better image quality and larger DMOS values indicate poorer image quality. The range of DMOS values is [0, 100] for the LIVE database, [0, 1] for the CSIQ database, [0, 9] for the MOS of the TID2013 database, [1, 5] for the DMOS of the KADID-10K database, [0, 100] for the MOS of the LIVEC database, and [1, 5] for the MOS of the KonIQ-10K database. The subjective quality score was scaled to [0,1] using the Min-Max Normalization, which can be formulated as:
\begin{equation}
    S = \frac{S-S_{min}}{S_{max}-S_{min}} 
\end{equation}
where $S$ denotes the subjective quality score.

\begin{table}\centering
\caption{Summary of IQA datasets.}\label{tab:1}
\label{tab:my-table}
\scalebox{1.0}{
\begin{tabular}{@{}cccl@{}}
\toprule
Databases & Dist. Images & Dist. types & DB. type  \\ \midrule
LIVE      & 799          & 5           & Synthetic \\
CSIQ      & 866          & 6           & Synthetic \\
TID2013   & 3000         & 24          & Synthetic \\
KADID-10K & 10125        & 25          & Synthetic \\
LIVEC     & 1162         & -           & Authentic \\
KonIQ-10k & 10073        & -           & Authentic \\ \bottomrule
\end{tabular}}
\end{table}

\subsection{Evaluation Metrics}
We use Spearman's rank order correlation coefficient (SROCC) and Pearson's linear correlation coefficient (PLCC) to measure the performance of the NR-IQA method. SROCC is defined as follows:

\begin{equation}
\mathrm{\textbf{SROCC}}=1-\frac{6 \sum_{i=1}^{n} d_{i}^{2}}{n\left(n^{2}-1\right)}
\end{equation}

\begin{table*}
\centering
\caption{Comparison of \textit{\textbf{TempQT}} \textit{v.s.} state-of-the-art NR-IQA algorithms on synthetically and authentically distorted datasets. Performance scores of other methods are as reported in the corresponding original papers. Best scores are $\textbf{bolded}$, second best are \underline{underlined}, missing scores are shown as $``–"$ dash.}\label{tab:2}
\scalebox{1.0}{
\begin{tabular}{c!{\vrule width \lightrulewidth}cc!{\vrule width \lightrulewidth}cc!{\vrule width \lightrulewidth}cc!{\vrule width \lightrulewidth}cc!{\vrule width \lightrulewidth}cc!{\vrule width \lightrulewidth}cc} 
\toprule
\multirow{2}{*}{} & \multicolumn{2}{c!{\vrule width \lightrulewidth}}{CSIQ} & \multicolumn{2}{c!{\vrule width \lightrulewidth}}{LIVE} & \multicolumn{2}{c!{\vrule width \lightrulewidth}}{LIVE challenge} & \multicolumn{2}{c!{\vrule width \lightrulewidth}}{TID2013} & \multicolumn{2}{c!{\vrule width \lightrulewidth}}{KonIQ-10k} & \multicolumn{2}{c}{Average}      \\
\cmidrule{2-13}
                  & SROCC          & PLCC                                   & SROCC          & PLCC                                   & SROCC          & PLCC                                             & SROCC          & PLCC                                      & SROCC          & PLCC                                        & SROCC          & PLCC            \\ 
\midrule
HFD*\cite{wu2017hierarchical}              & 0.842          & 0.890                                  & 0.951          & 0.971                                  & -              & -                                                & 0.764          & 0.681                                     & -              & -                                           & -              & -               \\
PQR*\cite{zeng2017probabilistic}              & 0.873          & 0.901                                  & 0.965          & 0.971                                  & 0.808          & 0.836                                            & 0.849          & 0.864                                     & -              & -                                           & -              & -               \\
DIIVINE\cite{saad2012blind}           & 0.804          & 0.776                                  & 0.892          & 0.908                                  & 0.588          & 0.591                                            & 0.643          & 0.567                                     & 0.546          & 0.558                                       & 0.695          & 0.680           \\
BRISQUE\cite{mittal2012no}           & 0.812          & 0.748                                  & 0.929          & 0.944                                  & 0.629          & 0.629                                            & 0.626          & 0.571                                     & 0.581          & 0.685                                       & 0.715          & 0.715           \\
ILNIQE\cite{zhang2015feature}            & 0.822          & 0.865                                  & 0.902          & 0.906                                  & 0.508          & 0.508                                            & 0.521          & 0.648                                     & 0.523          & 0.537                                       & 0.655          & 0.693           \\
BIECON\cite{kim2016fully}            & 0.815          & 0.823                                  & 0.958          & 0.961                                  & 0.613          & 0.613                                            & 0.717          & 0.762                                     & 0.651          & 0.654                                       & 0.751          & 0.763           \\
MEON\cite{ma2017end}              & 0.852          & 0.864                                  & 0.951          & 0.955                                  & 0.697          & 0.710                                            & 0.808          & 0.824                                     & 0.611          & 0.628                                       & 0.784          & 0.796           \\
WaDIQaM\cite{bosse2017deep}           & 0.852          & 0.844                                  & 0.960          & 0.955                                  & 0.682          & 0.671                                            & 0.835          & 0.855                                     & 0.804          & 0.807                                       & 0.827          & 0.826           \\
TIQA\cite{you2021transformer}              & 0.825          & 0.838                                  & 0.949          & 0.965                                  & 0.845          & 0.861                                            & 0.846          & 0.858                                     & 0.892          & 0.903                                       & 0.871          & 0.885           \\
MetaIQA\cite{zhu2020metaiqa}           & 0.899          & 0.908                                  & 0.960          & 0.959                                  & 0.802          & 0.835                                            & 0.856          & 0.868                                     & 0.887          & 0.856                                       & 0.881          & 0.885           \\
P2P-BM\cite{ying2020patches}            & 0.899          & 0.902                                  & 0.959          & 0.958                                  & 0.844          & 0.842                                            & 0.862          & 0.856                                     & 0.872          & 0.885                                       & 0.887          & 0.889           \\
HyperIQA\cite{su2020blindly}          & \underline{0.923}  & \underline{0.942}                          & 0.962          & 0.966                                  & \underline{0.859} & \underline{0.882}                                   & 0.840          & 0.858                                     & \underline{0.906}          & 0.917                                       & 0.898          & 0.913           \\
TReS\cite{golestaneh2022no}              & 0.922          & \underline{0.942}                          & \underline{0.969}          & \underline{0.968}                                  & 0.846          & 0.877                                    & \underline{0.863}  & \underline{0.883}                             & \textbf{0.915} & \textbf{0.928}                              & \underline{0.903}  & \underline{0.920}   \\ 
\midrule
TempQT          & \textbf{0.950} & \textbf{0.960}                         & \textbf{0.976} & \textbf{0.977}                         & \textbf{0.870}  & \textbf{0.886}                                            & \textbf{0.883} & \textbf{0.906}                            & 0.903          & \underline{0.920}                                       & \textbf{0.916} & \textbf{0.930}  \\
\toprule
\end{tabular}}
\end{table*}

\noindent where $n$ is the number of test images and $d_{i}$ denotes the difference between the ranks of $i$-th test image in ground-truth and the predicted quality scores. PLCC is defined as:

\begin{equation}
\mathrm{\textbf{PLCC}}=\frac{\sum_{i=1}^{n}\left(u_{i}-\bar{u}\right)\left(v_{i}-\bar{v}\right)}{\sqrt{\sum_{i=1}^{n}\left(u_{i}-\bar{u}\right)^{2}} \sqrt{\sum_{i=1}^{n}\left(v_{i}-\bar{v}\right)^{2}}}
\end{equation}

\noindent where $u_{i}$ and $v_{i}$ denote the ground-truth and predicted quality scores of the $i$-th image, and $\bar{u}$ and $\bar{v}$ are their mean values, respectively.

\subsection{Implementation Details}

We implemented our model by PyTorch and conducted training and testing on an NVIDIA RTX 3090 GPU. Following the standard training strategy from existing IQA algorithms, we randomly sampled and flipped 25 patches horizontally and vertically with the size of 224×224 pixels from each training image for augmentation. Training patches inherited quality scores from the source image. In the training stage, we perform pre-training of the Transformer encoder and decoder by minimizing $\mathcal{L}_{EM}$ on the KADID-10K training set; in the testing stage, we train the final quality model by minimizing $\mathcal{L}_{Q}$ on the training set. We used Adam \cite{kingma2014adam} optimizer with weight decay $1\times 10^{-5}$ to train our model for at most 15 epochs, with a mini-batch size of 16. The learning rate is first set to $2\times 10^{-5}$, and reduced 0.9 times the original rate after every 5 epochs. We use $L$ as the number of encoder layers in the Transformer, $D$=768, $p$=16, and set the number of heads $h$=16. Specifically, for each dataset the parameters may be slightly adjusted due to differences in resolution and dataset size.

Following the common practice in NR-IQA \cite{su2020blindly,golestaneh2022no}, all experiments use the same setting, where we first select 10 different seeds, and then use them to split the datasets randomly to train/test (80$\%$/20$\%$). So we have a total of 10 different splits. Testing data is not being used during the training. In the case of synthetically distorted datasets, the split is implemented according to reference images to avoid content overlapping. For the results of all experiments, we run the experiments 10 times with different initializations and report the average values of SROCC and PLCC.

\subsection{Performance Comparison}

Table \textcolor{red}{\ref{tab:2}} shows the overall performance comparison in terms of SROCC and PLCC on several standard image quality datasets, which cover both synthetically and authentically distorted images. Since KADID-10k is used as pre-training for OEM, it is not involved in the result comparison. Our model achieves the best results on all the  standard datasets except KonIQ-10k, where we achieve the second best for the PLCC and still competitive performance for SROCC.  In the last column, we provide the average performance across all datasets, and we observe that our proposed method outperforms existing methods on both SROCC and PLCC.

In Table \textcolor{red}{\ref{tab:3}}, we conduct cross dataset evaluations and compare our model to the competing approaches. Training is performed on one specific dataset, and testing is performed on another different dataset without any fine-tuning or parameter adaptation. As shown in Table \textcolor{red}{\ref{tab:3}}, our proposed method outperforms other algorithms on three out of four datasets, which indicate the strong generalization ability of our approach.

Since distortion types are diverse and generally unknown on authentic image databases and one image may contain multiple types of noises, to verify the generalization performance of our proposed model on specific distortion types, we compared the SROCC results on synthetic distortion databases LIVE and CSIQ. In Table \textcolor{red}{\ref{tab:4}}, it can be seen that our proposed model has better generalization performance on LIVE for White Noise and Gaussian Blur distortions.  In the CSIQ database, our models outperforms the current state-of-the-art in White Noise, JPEG, JPEG2000, FNoise, Gaussian Blur and Contrast distortion types. This also proves that our proposed model has better generalization performance and can be employed to evaluate the image quality degradation based on an understanding of the image content.

\begin{table}
\centering
\caption{SROCC evaluations on cross datasets, where \textbf{bold} indicate the best performers, and second best are \underline{underlined}.}\label{tab:3}
\begin{tabular}{c|l!{\vrule width \lightrulewidth}c|cc} 
\toprule
\multicolumn{1}{c!{\vrule width \lightrulewidth}}{Trained on} & KonIQ          & LIVEC          & \multicolumn{2}{c}{LIVE}         \\ 
\midrule
Test on                                                       & LIVEC          & KonIQ          & CSIQ           & TID2013         \\ 
\midrule
WaDIQaM\cite{bosse2017deep}                                                       & 0.682          & 0.711          & 0.704          & 0.462           \\
P2P-BM\cite{ying2020patches}                                                        & 0.770          & 0.740          & 0.712          & 0.488           \\
HyperIQA\cite{su2020blindly}                                                      & 0.785          & \textbf{0.772} & 0.744          & 0.551           \\
Tres\cite{golestaneh2022no}                                                          & \underline{0.786}          & 0.733          & \underline{0.761}  & \underline{0.562}   \\ 
\midrule
\multicolumn{1}{c!{\vrule width \lightrulewidth}}{TempQT}   & \textbf{0.789} & \underline{0.750}  & \textbf{0.821} & \textbf{0.575}  \\
\toprule
\end{tabular}
\end{table}
\subsection{Visualization}

\begin{figure}
    \centering
    \includegraphics[scale=0.115]{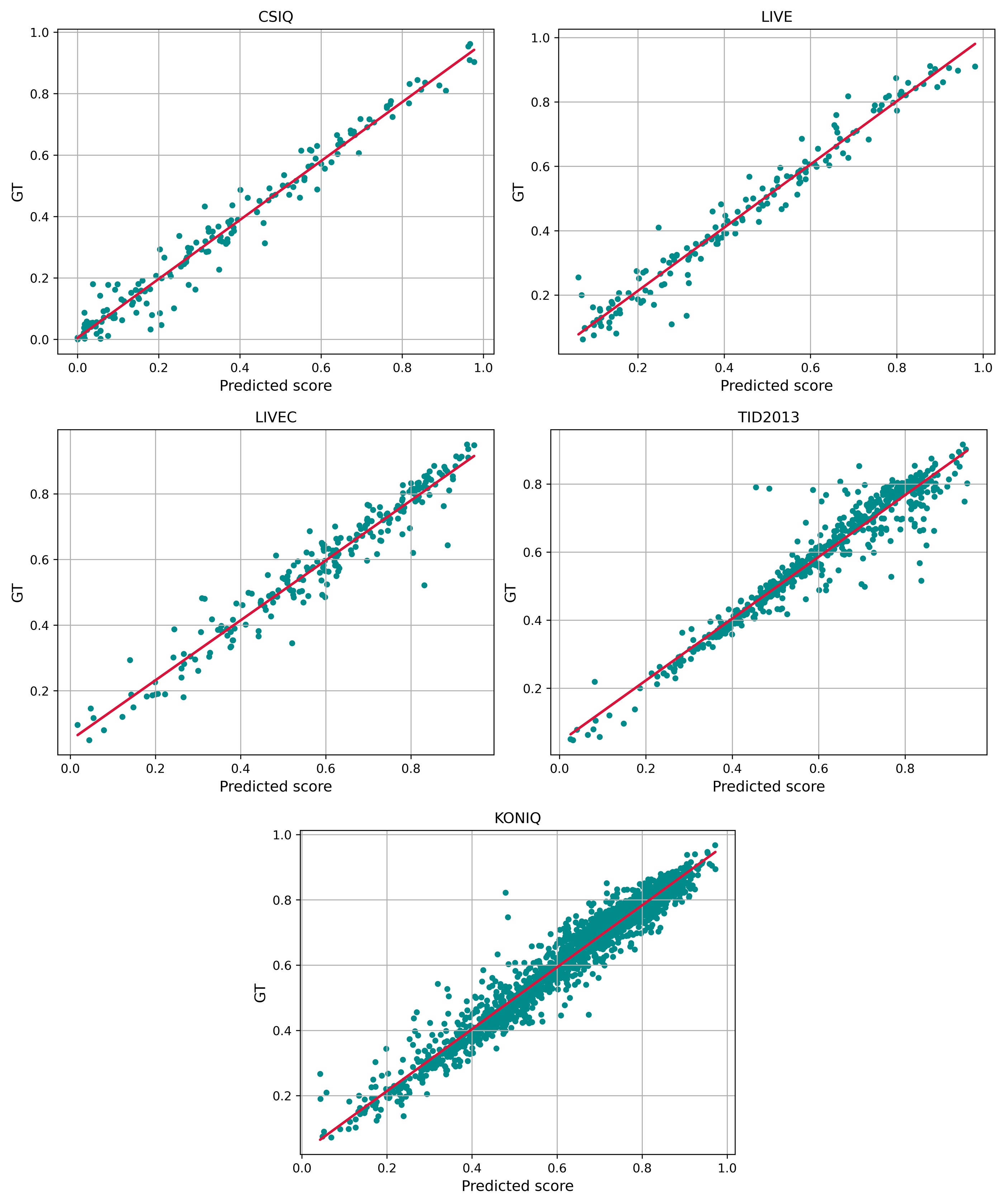}
    \caption{Scatter plots of ground-truth against predicted scores of proposed \textit{\textit{TempQT}} on CSIQ, LIVE, LIVE challenge, TID2013 and KonIQ datasets.}
    \label{fig:fig3}
\end{figure}

Figure \textcolor{red}{\ref{fig:fig3}} shows the scatter plots for the subjective scores of distorted images and model-predicted values on the test set, where the red straight lines indicate the linear fitting function. On the CSIQ, LIVE and TID2013 datasets, our model predicts the quality scores very well with a strong linear relationship between the predicted values and GT, and combing with the results in Table \textcolor{red}{\ref{tab:2}}, the model is ranked the first in both SROCC and PLCC values. On the LIVE challenge dataset, the model's predictions also has a strong linear relationship with GT, where our model is also ranked the first in this dataset for SROCC  and PLCC values. This shows that the prediction of our proposed model is very effective and accurate.

\begin{table*}
\centering
\caption{SROCC comparisons on individual distortion types on the LIVE and CSIQ databases, where \textit{bold} indicate the best performers}\label{tab:4}
\begin{tabular}{c!{\vrule width \lightrulewidth}ccccc!{\vrule width \lightrulewidth}cccccc} 
\toprule
Databse  & \multicolumn{5}{c}{LIVE}                                                           & \multicolumn{6}{c}{CSIQ}                                                                                                           \\ 
\midrule
Type     & JP2K           & JPEG           & WN             & GB             & FF             & WN             & JPEG           & JP2K           & FN             & GB             & \begin{tabular}[c]{@{}c@{}}CC\\\end{tabular}  \\ 
\toprule
BRISQUE\cite{mittal2012no}  & 0.929          & 0.965          & 0.982          & 0.964          & 0.828          & 0.723          & 0.806          & 0.840          & 0.378          & 0.820          & 0.804                                         \\
ILNIQE\cite{zhang2015feature}   & 0.894          & 0.941          & 0.981          & 0.915          & 0.833          & 0.850          & 0.899          & 0.906          & 0.874          & 0.858          & 0.501                                         \\
HOSA\cite{xu2016blind}     & 0.935          & 0.954          & 0.975          & 0.954          & \textbf{0.954} & 0.604          & 0.733          & 0.818          & 0.500          & 0.841          & 0.716                                         \\
BIECON\cite{kim2016fully}   & 0.952          & \textbf{0.974} & 0.980          & 0.956          & 0.923          & 0.902          & 0.942          & 0.954          & 0.884          & 0.946          & 0.523                                         \\
WaDIQaM\cite{bosse2017deep}  & 0.942          & 0.953          & 0.982          & 0.938          & 0.923          & 0.974          & 0.853          & 0.947          & 0.882          & 0.976 & 0.923                                         \\
PQR\cite{zeng2017probabilistic}      & \textbf{0.953} & 0.965          & 0.981          & 0.944          & 0.921          & 0.915          & 0.934          & 0.955          & 0.926          & 0.921          & 0.837                                         \\
HyperIQA\cite{su2020blindly} & 0.949          & 0.961          & 0.982          & 0.926          & 0.934          & 0.927          & 0.934          & 0.960          & 0.931          & 0.915          & 0.874                                         \\ 
\midrule
TempQT     & 0.929          & 0.944          & \textbf{0.988} & \textbf{0.987} & 0.945          & \textbf{0.987} & \textbf{0.987} & \textbf{0.985} & \textbf{0.987} & \textbf{0.978}          & \textbf{0.966}                                \\
\cmidrule[\heavyrulewidth]{1-1}\cmidrule[\heavyrulewidth]{2-12}
\end{tabular}
\end{table*}

\begin{figure}
    \centering
    \includegraphics[scale=0.45]{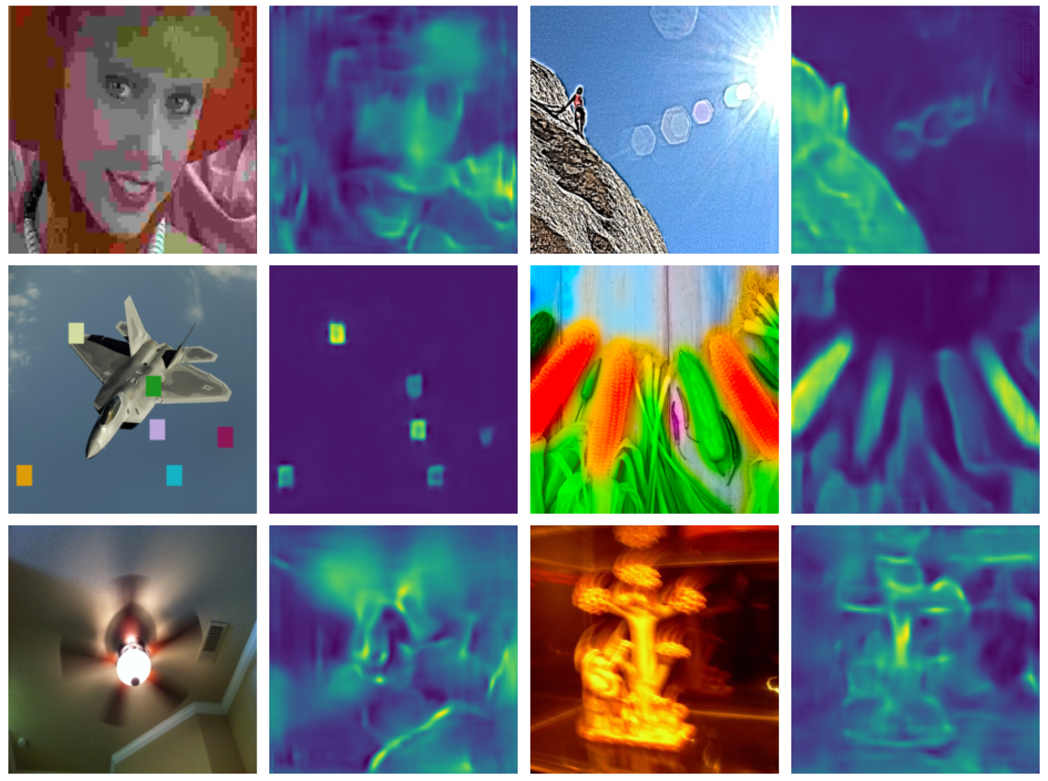}
    \caption{PEMs generated using our proposed model. In each pair, the left denotes the distorted image, and the right denotes the error maps blended with the originals using viridis color.}
    \label{fig:fig4}
\end{figure}

In Figure \textcolor{red}{\ref{fig:fig4}}, we show the error map extracted by the pre-trained model. The distorted images are selected from the KADID-10K and LIVE challenge datasets, and the bright part in the map indicates the distorted areas extracted by the pre-trained model, and the brighter the distortion is, the more severe the distortion is. From these error maps, it is clear that our model can effectively capture various distortion information such as blur, motion blur, over-saturation and noisy color blocks.

\begin{figure}
    \centering
    \includegraphics[scale=0.45]{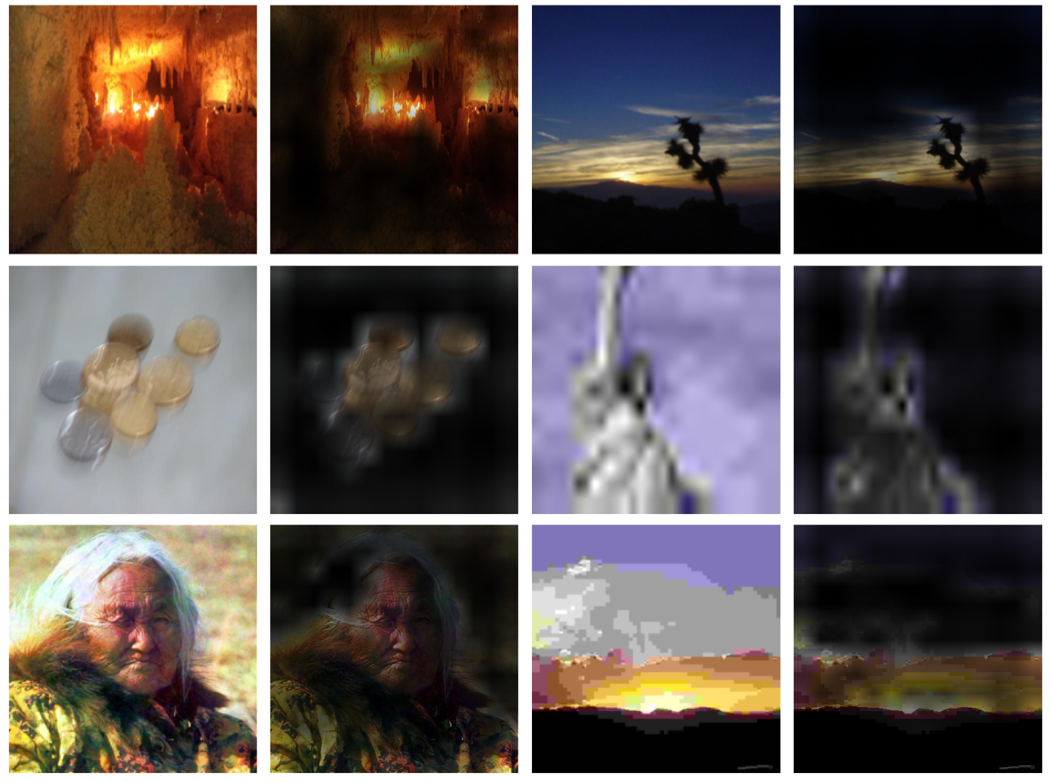}
    \caption{Visualization of Attention Maps (AM) from the proposed TempQT. The images are randomly sampled from the LIVE and CSIQ datasets. The left denotes the distorted image, and the right denotes the AM output.}
    \label{fig:fig5}
\end{figure}

In Figure \textcolor{red}{\ref{fig:fig5}}, we show the Attention Map (AM) extracted by the TempQT model. AM indicates the region that the model is most concerned with in predicting the image quality. Since our model only uses the encoder part of the Transformer, Am actually comes from attention vector of the perceptual quality token in the MHSA layer. As outlined in Section III-B, We first mapped it by averaging the attention maps of all layers and then resized the attention maps to the distorted image size. When performing quality evaluation, it is important to combine the global information to evaluate the local distortion of the  images. From the figure, we can see that the attention of our model is more evenly distributed in the distortion-information concentrated part in the distorted image, which is in line with the evaluation behaviours of human eyes. Our proposed perceptual quality-based on Transformer and PEM is more effective in performing  evaluation for distorted image.

\subsection{Ablation Study and Discussion}

In Table \textcolor{red}{\ref{tab:5}}, we provide ablation experiments to illustrate the effect of each component of our proposed method by comparing the results on LIVE challenge, LIVE and CSIQ datasets. It can be seen that neither the PEM branch nor the PQT branch alone is very effective for image quality score prediction, and only a combination of the two is able to achieve the best prediction performance. 

\begin{figure}
    \centering
    \includegraphics[scale=0.5]{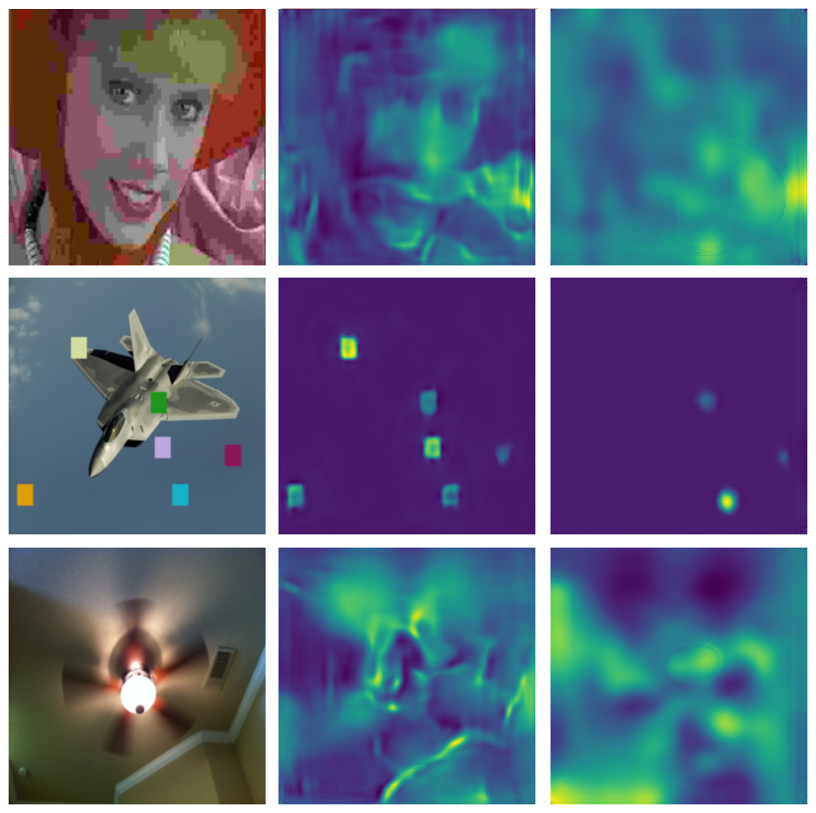}
    \caption{PEMs comparison between multi-layer and last layer. The left denotes the distorted image, the middle denotes the multi-layer output, and the right denotes the last layer output.}
    \label{fig:fig6}
\end{figure}

During the training of the two-branch Transformer model, we found that if the Transformer model parameters of the PEM branch were migrated to the PQT branch Transformer, the prediction of the final image quality score would be worse. Considering that PEM is mainly concerned with image distortion information and PQT is more concerned with the perceptual effect of the distortion information in an image, the overlapping of the two is not very much. They can even be seen as two different tasks. Therefore, though directly sharing model parameters will reduce the complexity of the model, it will lead to the degradation of model prediction performance. The comparison results for the  model parameter sharing are shown in Table \textcolor{red}{\ref{tab:6}}. 
In addition, how to select Transformer layers for generating the PEM also has an impact on the final result. In IQA, as a visual task that favors low-level details of an image, a combination of shallow low-level and deep high-level information of Transformer is often more capable of representing the structural and semantic information of an image's details. This will result in better overall evaluation performance of the final distorted image, as shown in Table \textcolor{red}{\ref{tab:7}}. Furthermore, for visualization purpose, in Figure \textcolor{red}{\ref{fig:fig6}}, we use multi-layer ([0, 2, 6 ,11]) and last layer ([11]) to generate PEM respectively. Although each layer of the Transformer outputs the same resolution of the feature map and has semantic-level information, the low-level distortion details about the image are essential in generating an effective PEM for the IQA task, which  explains why the multi-layer approach works better.

\begin{table}
\centering
\caption{Ablation experiments on the effects of different components for our proposed model. PQT denotes Perceptual Quality Token.}\label{tab:5}
\scalebox{0.9}{
\begin{tabular}{cc!{\vrule width \lightrulewidth}cc!{\vrule width \lightrulewidth}cc!{\vrule width \lightrulewidth}cc} 
\toprule
\multirow{2}{*}{PEM} & \multirow{2}{*}{PQT} & \multicolumn{2}{c!{\vrule width \lightrulewidth}}{LIVE challenge} & \multicolumn{2}{c!{\vrule width \lightrulewidth}}{LIVE} & \multicolumn{2}{c}{CSIQ}  \\ 
\cmidrule{3-8}
                     &                     & SROCC & PLCC                                             & SROCC & PLCC                                            & SROCC & PLCC              \\ 
\midrule
\checkmark                    &                     & 0.804 & 0.817                                            & 0.956 & 0.957                                           & 0.902 & 0.905             \\
                     & \checkmark                    & 0.823 & 0.860                                            & 0.954 & 0.955                                           & 0.914 & 0.916             \\
\checkmark                     & \checkmark                    & 0.870 & 0.886                                            & 0.976 & 0.977                                           & 0.950 & 0.960             \\
\toprule
\end{tabular}}
\end{table}

\begin{table}
\centering
\caption{Comparison of shared model parameters between SROCC and PLCC on LIVE, CSIQ, and LIVE challenge databases, where PS denotesparameter sharing}\label{tab:6}
\scalebox{0.9}{
\begin{tabular}{c!{\vrule width \lightrulewidth}cc!{\vrule width \lightrulewidth}cc!{\vrule width \lightrulewidth}cc} 
\toprule
\multirow{2}{*}{} & \multicolumn{2}{c!{\vrule width \lightrulewidth}}{LIVE} & \multicolumn{2}{c!{\vrule width \lightrulewidth}}{CSIQ} & \multicolumn{2}{c}{LIVE challenge}  \\ 
\cmidrule{2-7}
                  & SROCC & PLCC                                            & SROCC & PLCC                                            & SROCC & PLCC                        \\ 
\midrule
w/ PS            & 0.920 & 0.927                                           & 0.927 & 0.948                                           & 0.820 & 0.825                       \\
w/o PS          & 0.976 & 0.977                                           & 0.950 & 0.960                                           & 0.870 & 0.886                       \\
\toprule
\end{tabular}}
\end{table}

\begin{table}
\centering
\caption{Comparison of SROCC and PLCC with different selected layers on CSIQ and LIVE challenge databases}\label{tab:7}
\scalebox{0.9}{
\begin{tabular}{c!{\vrule width \lightrulewidth}cc!{\vrule width \lightrulewidth}cc} 
\toprule
\multirow{2}{*}{Selected layers} & \multicolumn{2}{c!{\vrule width \lightrulewidth}}{CSIQ} & \multicolumn{2}{c}{LIVE challenge}  \\ 
\cmidrule{2-5}
                                 & SROCC & PLCC                                            & SROCC & PLCC                        \\ 
\midrule
{[}11]                           & 0.940 & 0.953                                           & 0.830 & 0.874                       \\
{[}1, 3, 5, 7, 9, 11]            & 0.929 & 0.944                                           & 0.836 & 0.876                       \\
{[}0, 2, 6, 11]                  & 0.950 & 0.960                                           & 0.870 & 0.886                       \\
\toprule
\end{tabular}}
\end{table}

\section{Conclusions}
In this paper, we propose a new NR-IQA algorithm based on Predicted Error Map (PEM) and Perceptual Quality Token (PQT) using vision Transformer. Firstly, we obtain the PEM by pre-training the Transformer model, and then we fuse the PEM with PQT for feature aggregation. Finally, we use the fused features for blind quality assessment of distorted images. Our experiments show that our proposed method outperforms  the current state-of-the-art on both synthetic and authentic IQA datasets. In addition, experiments on the cross dataset and individual distortion types also reveal that the model evaluates the unknown noise-distorted images with accurate results, and thus our proposed model has better generalization performance.
More visualization results about objective error map and perceptual attention map are provided in supplementary material.


\bibliographystyle{IEEEtran_bib}
\bibliography{ref}

\end{document}